\documentclass[12pt]{amsart}
\usepackage{geometry}               
\usepackage{amssymb,amsmath,amsthm}
\usepackage{xcolor}
\definecolor{myurlcolor}{rgb}{0,0,0.4}
\definecolor{mycitecolor}{rgb}{0,0.5,0}
\definecolor{myrefcolor}{rgb}{0.5,0,0}
\usepackage[pagebackref,colorlinks,linkcolor=myrefcolor,citecolor=mycitecolor,urlcolor=myurlcolor,hypertexnames=true]{hyperref}
\usepackage{amsrefs}
\usepackage{graphicx}
\usepackage{enumitem}
\usepackage[font=scriptsize]{caption}
\usepackage[font=footnotesize]{subcaption}

\usepackage[utf8]{inputenc}
\usepackage{float}

\theoremstyle{plain}
\newtheorem{theorem}{Theorem}[section]

\newtheorem{algorithm}[theorem]{Algorithm}

\theoremstyle{remark}

\newtheorem{remark}[theorem]{Remark}

\theoremstyle{definition}

\def\R{\mathbb{R}}

\newcommand{\norm}[1]{\left\lVert #1 \right\rVert}

\newcommand{\mcal}[1]{\mathcal{#1}}
\newcommand{\pc}{\mathcal{X}}

\newcommand{\featdim}{d} % dim of feature space
\newcommand{\encoder}{h}
\newcommand{\decoder}{g}
\newcommand{\autoencoder}{f}
\newcommand{\pcy}{\mathcal{Y}}
\newcommand{\Chamfer}{\operatorname{Chamfer}}
\newcommand{\EMD}{\operatorname{EMD}}
\newcommand{\Loss}{\operatorname{L}}

\let\originalleft\left % improve 
\let\originalright\right % spacing
\renewcommand{\left}{\mathopen{}\mathclose\bgroup\originalleft} % around
\renewcommand{\right}{\aftergroup\egroup\originalright} % \left and \right

\title[Cluster-Based AEs for Volumetric Point Clouds]{Cluster-Based Autoencoders for Volumetric Point Clouds}

\author[S. Antholzer]{Stephan Antholzer}
\address{Stephan Antholzer, Universit\"at Innsbruck, Austria }
\email{stephan.antholzer@uibk.ac.at}

\author[M. Berger]{Martin Berger}
\address{Martin Berger, Universit\"at Innsbruck, Austria }
\email{martin.berger@uibk.ac.at}

\author[T. Hell]{Tobias Hell}
\thanks{Corresponding author: \href{mailto:tobias.hell@uibk.ac.at}{tobias.hell@uibk.ac.at}}
\address{Tobias Hell, Universit\"at Innsbruck, Austria}
\email{tobias.hell@uibk.ac.at}

\begin{document}

\begin{abstract} 
Autoencoders allow to reconstruct a given input from a small set of parameters. However, the input size is often limited due to computational costs. We therefore propose a clustering and reassembling method for volumetric point clouds, in order to allow high resolution data as input. We furthermore present an autoencoder based on the well-known FoldingNet \cite{foldingnet} for volumetric point clouds and discuss how our approach can be utilized for blending between high resolution point clouds as well as for transferring a volumetric design/style onto a pointcloud while maintaining its shape.
\end{abstract}

\maketitle

\section{Introduction}
\label{sec: introduction}

\begin{figure}[b]
\centering
\includegraphics[scale=.55]{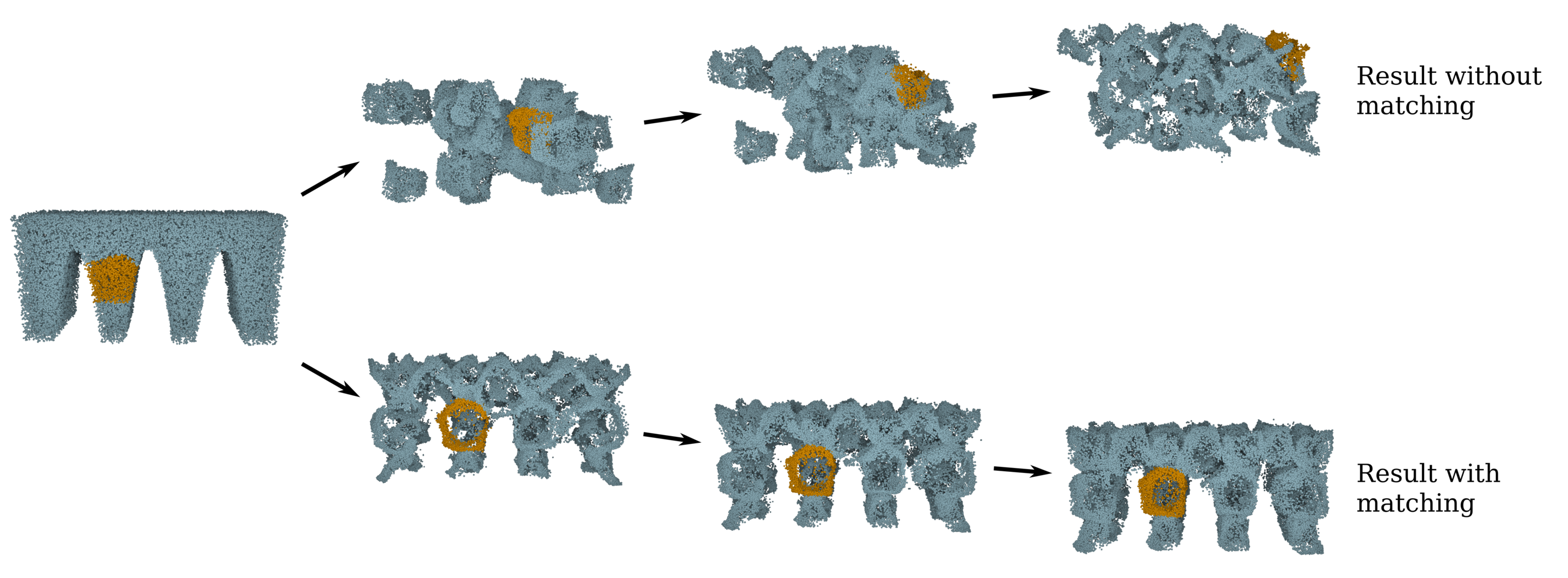}
\caption{Applying the stripes design onto a point cloud representing a bridge. The upper row shows the blending process with mismatched clusters. The lower row shows the blending process when using our proposed method in Section \ref{subsec: generating}.}\label{fig::clustercomparison}
\end{figure}

Point clouds are a versatile way to represent geometric structures, gaining more and more popularity in the fields of computer vision, see for instance \cites{stein, boulch, golovinsky, hackel, pointnet} and architecture \cites{wir, mathias, architectureAI1, architectureAI2}. However, their irregular structure causes new challenges within deep neural network frameworks. Therefore, point clouds are often transformed into rigid voxel grids before being used as input for neural networks, see e.g.\ \cites{wu, voxnet}.

In this paper, we continue the study of point clouds from the viewpoint of neural networks. We disscuss a cluster and reassembling method, which allows for high resolution inputs for any autoencoder for point clouds, while still being able to blend between different input data within the feature space. We then propose an autoencoder for volumetric point clouds and offer a method for transferring a volumetric design/style onto a point cloud while maintaining its shape. Our contribution is as follows. 

In Section 2 we start with a discussion of a \textit{constrained} $k$\textit{-means clustering} algorithm based on \cite{clustering} for the purpose of clustering a given point cloud into clusters of the same size. These clusters will then be used as input for an autoencoder. We then propose a reassembling method with the aim of still allowing to blend between two point clouds within the feature space of the autoencoder. Combining these methods, allows us to apply standard autoencoder procedures to high resolution point clouds. 

We continue by introducing an autoencoder based on~\cite{foldingnet} which is able to encode/decode volumetric point clouds. This is achieved by using a weighted sum of the well-known Chamfer loss and the Earth Mover's distance which penalizes discrepancies in shape as well as interior structures. We then proceed with a method which allows us to apply a design style represented by a \textit{design point cloud} onto a given point cloud $\pc$ which represents a geometrical object. The idea is to subsample the design point cloud according to the probability distribution underlying the point cloud $\pc$, and then use the autoencoder to restore potentially lost geometric properties of $\pc$ by blending within the feature space of the autoencoder between the original point cloud $\pc$ and the subsampled point cloud.
We conclude Section 2 by giving two pipelines on how to combine the previously discussed methods in order to generate new point clouds. More particularly, we discuss how to properly use the clusters obtained by applying the aforementioned clustering algorithm. While blending within the feature space between two inputs of an autoencoder is usually a routine method, in this case it is problematic when combined with prior clustering and reassembling. More specifically, one needs to assure that the blending occurs between corresponding clusters.
We therefore suggest to only cluster one point cloud and then use the corresponding clusters as inputs for a cluster assignment task regarding the second point cloud. This approach assures a proper reassembling of the clusters of the point cloud obtained through blending.

We continue in Section 3 with a description of our training dataset. While the autoencoder in \cite{foldingnet} was originally trained on the Shapenet dataset \cite{shapenet}, we used a new collection of volumetric point clouds which allows the autoencoder to recognize shapes as well as interior structures. Essentially our training set consists of four different types of point clouds, encapsulated spheres, encapsulated cuboids, orthogonally intersecting planes and lattices with various geometric shapes.  This dataset is rich enough in structure in order for the autoencoder to learn how to reconstruct volumetric point clouds.

In Section 4, we describe our experimental setup as well as results. Our experiments show the effectiveness of our suggested methods. Due to the clustering, the autoencoder is able to reconstruct even fragile structures within high resolution pointclouds and by applying our style transfer method, we maintain the geometry of the underlying ground truth but impose the desired volumetric design onto them.

\section{Methodology}
\label{sec: Methodology}
Throughout, let $n$ denote a positive integer and $\pc=\{x_1,\ldots,x_n\}\subseteq\R^3$ a finite set of points.

\subsection{Clustering}
\label{subsec: clustering}
When it comes to real world applications, point clouds of high resolution are often considered. However, the practicality of autoencoders for point clouds is often limited to rather small point clouds. This also applies to the the discussed autoencoder in the next Subsection \ref{subsec: autoencoder}. In order to make autoencoders nevertheless still applicable, the idea is to cluster the point clouds into smaller pieces and encode/decode these smaller clusters and reassemble them appropriately when reconstructing the point cloud. Note however, that typically the number of points $m$, which is fed into an autoencoder is fixed. Thus each cluster must contain the same amount of points. We therefore use a special case of a two step \textit{constrained $k$-means clustering} algorithm proposed in \cite{clustering}. 

\begin{algorithm}\label{alg: cluster}
Let $C_1^t,\ldots,C_k^t\in\R^3$ denote cluster centers at iteration $t$ where $k$ is a natural number such that $n/k=m$ is again a positive integer. Now the algorithm at step $t+1$ works as follows.
\begin{enumerate}[label=\arabic*.]
\item \textbf{Cluster Assignment.} Fix $C_1^t,\ldots,C_k^t$ and compute a solution $T_{i,h}^t$ of the following linear program:
\begin{align*}
\operatorname*{minimize}_{T_{i,h}^t} & \qquad \sum_{i=1}^n\sum_{h=1}^k T_{i,h}^t \left(\frac{1}{2}\Vert x_i-C_h^t\vert_2^2\right)\\
\text{subject to}& \qquad \sum_{i=1}^n T_{i,h}=m\quad\text{for}\quad h=1,\ldots,k,\\
 &\qquad  \sum_{h=0}^k T_{i,h}=1\quad\text{for}\quad i=1,\ldots,n,\\
& \qquad T_{i,h}\geq 0 \quad\text{for}\quad i=1,\ldots,n\text{ and } h=1,\ldots,k.\\
\end{align*}
\item \textbf{Cluster Update.} Update $C_h^{t+1}$ as follows:
\begin{align*}
C_h^{t+1} = 
\begin{cases}
\frac{\sum_{i=1}^n T_{i,h}^t x_i}{\sum_{i=1}^n T_{i,h}^t}\quad\text{if }\sum_{i=1}^n T_{i,h}^t>0,\\
C_h^t\quad\text{otherwise}.
\end{cases}
\end{align*}
\end{enumerate}
Stop when $C_h^{t+1}=C_h^t$ for every $h=1,\ldots, k$, else increment $t$ by 1 and go to the first step.
\end{algorithm}

\begin{remark}
By Proposition 2.3 in \cite{clustering}, Algorithm \ref{alg: cluster} terminates after a finite number of iterations at a cluster assignment that is locally optimal. Furthermore the cluster assignment in the above algorithm can be interpreted as a Minimum Cost Flow linear network optimization problem, see \cite{bertsekas} and again \cite{clustering}, and thus by Proposition 5.6 in \cite{bertsekas} it follows that there exists an optimal solution of the above Cluster Assignment problem such that $T_{i,h}\in\{0, 1\}$ for every $i, h$.
\end{remark}

\subsection{Auotencoder for Volumetric Point Clouds}
\label{subsec: autoencoder}
In this section, we describe the autoencoder that we use for our approach and provide a capable tool in order to handle volumetric point clouds. Our neural network is based on the FoldingNet autoencoder for point clouds~\cite{foldingnet}. The FoldingNet consists of perceptron layers~\cite{pointnet} and graph 
layers~\cite{shen2018mining} which are special types of layers for operating on point clouds that 
were successfully used for point cloud classification problems.
We can write the autoencoder as $\autoencoder = \decoder \circ \encoder$, where $\encoder\colon \R^{3n}\to \R^\featdim$ is the encoder and $\decoder\colon \R^\featdim \to \R^{3m}$ is the deocoder.
For the FoldingNet, the dimension of the features space $\featdim$ is fixed and in order to efficiently train it the number of input points $n$ and the number of output points $m$ are also fixed, but not necessarily coincide. Another hyperparameter that needs to be chosen is the number of nearest neighbours in the local covariance layer $k$, for more details see~\cite{foldingnet}.

In order to train an autoencoder on point clouds, two losses are commonly used the 
Chamfer distance \cite{fan2017point} and the Earth Mover's distance (EMD)~\cites{rubner2000earth, peyre2019computational}.
Let $\pc,\pcy \subset \R^3$ be two point clouds. The Chamfer distance is then given as
\begin{equation}\label{eq:chamfer}
    \Chamfer(\pc,\pcy) = \sum_{x\in\pc}\min_{y\in\pcy}\norm{x-y}_2^2 + \sum_{y\in\pcy}\min_{x\in\pc}\norm{x-y}_2^2\,.
\end{equation}

The EMD on the other hand can be interpreted as the minimal cost (in terms of distances) of 
converting one point cloud into another. For more details see e.g. \cites{rubner2000earth,geomlossphd}. In 
the case where $\pc$ and $\pcy$ have the same number of point, which will be the case for our experiments, we can write it as 
\begin{equation}\label{eq:emd}
    \EMD(\pc,\pcy) = \min_{\varphi \in \Phi}\sum_{x\in \pc}\norm{x - \varphi(x)}_2\,,
\end{equation}
where $\Phi=\{\varphi\colon \pc \to \pcy \mid \varphi \text{ bijectiv}\}$.
For a definition with point clouds of unequal size see e.g. \cite[Equation (3.166)]{geomlossphd}.
Note that this is equivalent to the Wasserstein-1-distance since the point clouds can be interpreted as uniform discrete measures in $\R^3$.

\begin{figure}[htbp]
\centering
\begin{subfigure}{.45\textwidth}
\centering
\includegraphics[scale=.18,trim = 400 0 600 0, clip]{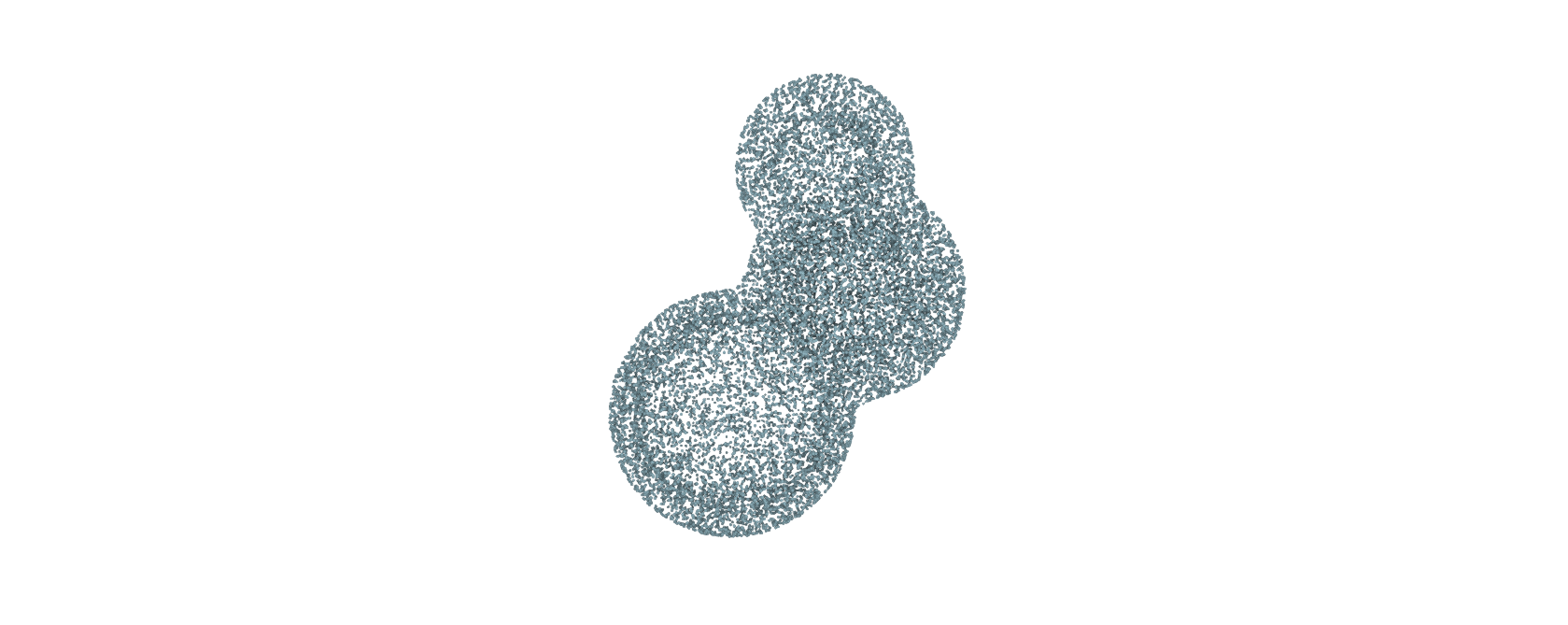}
\end{subfigure}
\hspace{-50pt}
\begin{subfigure}{.45\textwidth}
\centering
\includegraphics[scale=.18,trim = 500 0 600 0, clip]{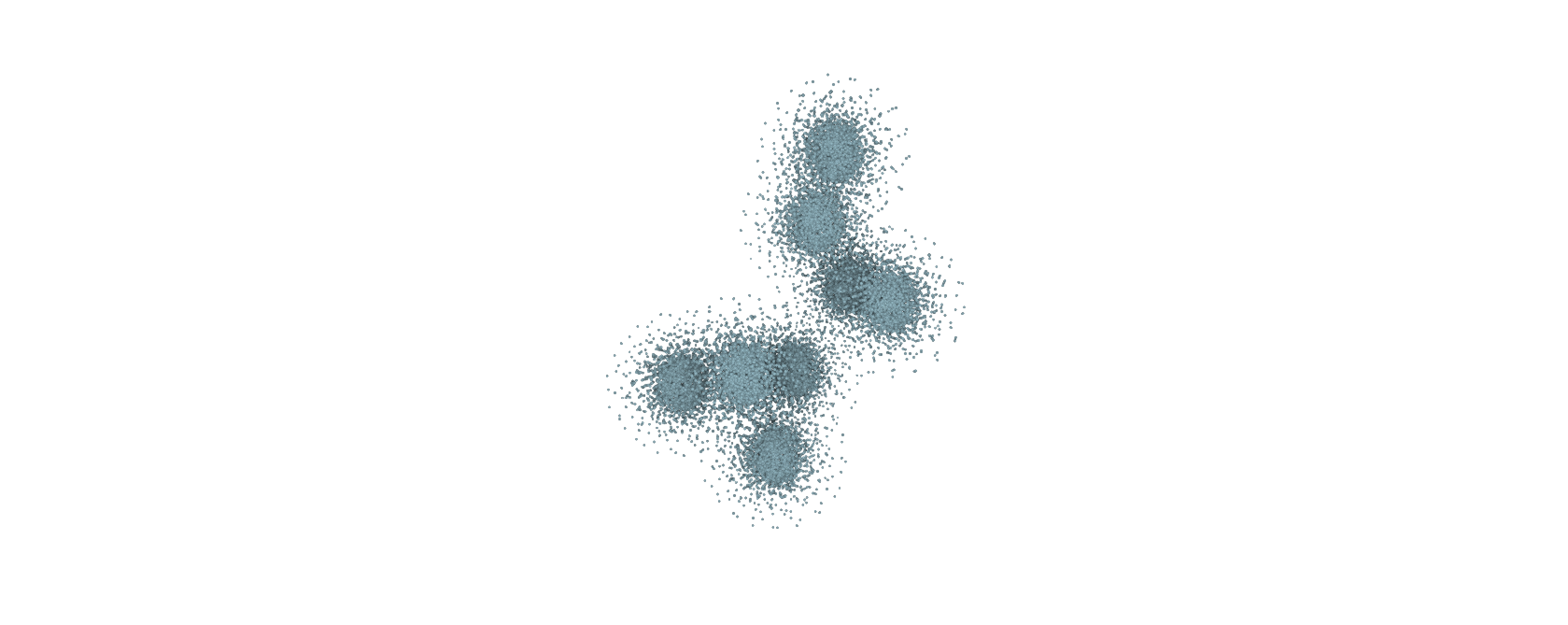}
\end{subfigure}
\caption{Reconstruction of a point cloud representing encapsulated spheres when only using the Chamfer distance as loss function. Left shows the ground truth, right the reconstruction.}\label{fig::onlychamfer}
\end{figure}

Since the Chamfer distance only minimizes minimal distances between the two point clouds it is not an 
ideal loss function for volumetric point clouds in our experiments. When training for instance
only with the Chamfer distance as a loss function, the resulting reconstructions did not caputure the 
structure of the point clouds, see Figure \ref{fig::onlychamfer}, and similar when only using the Earth Mover's distance.
The EMD on the other hand is a better fit for our kind of point clouds which results in the 
trained autoencoder being able to capture more details of the point clouds including structures on 
the inside of the cloud not only an outer shell.
However, calculating the EMD is very costly. In our implementation we did not use the exact EMD but 
the Sinkhorn divergence which is an approximation~\cites{peyre2019computational,geomloss,geomlossphd}.
The calculation times are still slower than the Chamfer distance if one wants to get a decent 
approximation.

We therefore propose that a combined loss function to capture the complex structure 
of our point clouds and which yields an autoencoder with desired properties. Thus our loss function $\Loss$ has the following form:

\begin{equation}\label{eq:loss}
    \Loss(\pc,\pcy) = \alpha_{\EMD} \EMD(\pc,\pcy) + \alpha_{\Chamfer} \Chamfer(\pc, \pcy)\,.
\end{equation}

Calculating the loss function becomes a major reason to limit the size of the point clouds during training. While it is possible to train with larger point clouds of e.g. 16384 points, the resulting time to optimize the networks would be too long and the number of points still to low.
This motivates our clustering approach outlined in the next Subsection~\ref{subsec: clustering}.

\subsection{Density Based Style Transfer}
\label{subsec: styletransfer}
Given a finite point cloud which represents some geometrical object, we want to apply a certain design or style onto it. In order to do so, we propose a density based style transfer approach which combines density estimates with the autoencoder $f=g\circ h$ introduced in Subsection \ref{subsec: autoencoder}, where $h$ is the encoder and $g$ is the decoder.
Let again $\pc$ denote a given finite point cloud in $\R^3$ with $n$ points. By scaling we may assume that $\pc$ lies within the cube $[0,1]^3$. Additionally let $\mcal{S}\subseteq [0, 1]^3$ also denote a finite \textit{design} point cloud within the unit cube, which is in some sense space filling, see for instance Figure \ref{fig: designcubes}. Before we proceed, note that a precise definition of the desired properties of $\mcal{S}$ highly depends on the use case and the goal of the design process. We thus intentionally leave the definition of $\mcal{S}$ vague.
Now assume that $\pc$ was sampled from some manifold according to some distribution with density function $f_{\pc}$. The idea of the style transfer is the following. We first sample according to the density $f_{\pc}$ a point cloud $\mcal{S}_{\pc}$ from $\mcal{S}$. Then we encode $\pc$ as well as $\mcal{S}_{\mcal{X}}$ and blend between the point clouds within the feature space, more particularly, let $\lambda\in [0,1]$ and compute
\begin{equation}\label{eq: styletransfer}
\pc_{\mcal{S}}=g\big(\lambda h(\pc)+(1-\lambda) h(\mcal{S}_{\pc})\big).
\end{equation}
Depending on the parameter $\lambda$ the new point cloud $\pc_{\mcal{S}}$ will then represent a new point cloud which maintains the main geometrical properties of $\pc$ while transferring the style of $\mcal{S}$ to its shape as well as interior structure. 
Note, that one could argue that the style transfer could be achieved by only sampling according to the density $f_{\pc}$. However, in general this approach will lose some properties of $\pc$ like its shape, due to voids within $\mcal{S}$, see for example the right plots in Figure \ref{fig::transfer_stripes}, \ref{fig::transfer_porous} and \ref{fig::transfer_cut}, especially in the pillar case, while the interpolation within the feature space and subsequent decoding restores these properties proportionally more with a larger $\lambda$.

\subsection{Generating new point clouds}
\label{subsec: generating}
Having examined the three essential building blocks of our methods, we can now discuss two different pipelines for generating new point clouds. Both methods heavily rely on blending between two point clouds within the feature space of the autoencoder.
However, this procedure can be problematic when it is applied to clusters instead of the whole point cloud.
Particularly, one needs to assure that the blending process takes place between corresponding clusters. We achieve this by only clustering one point cloud according to Algorithm \ref{alg: cluster} and then use the obtained centroids for a cluster assignment regarding the second point cloud. This yields the following two methods:

\begin{enumerate}[label=(\Roman*)]
\item \label{pipe: blending}\textit{Blending between two point clouds}. Let $\pc$ and $\mcal{Y}$ denote finite point clouds in $\R^3$ with the same number of points. Furthermore assume they are normalized such that $\pc,\mcal{Y}\subseteq [0,1]^3$.
Our goal is to create a new point cloud by mixing the geometric properties of $\pc$ and $\mcal{Y}$. First cluster $\pc$ according to Algorithm \ref{alg: cluster}, resulting in the clusters $\pc_1,\ldots, \pc_k$ with centroids $C_1,\ldots, C_k$ for some integer $k\geq 1$.
We then cluster $\mcal{Y}$ by applying the \textit{Cluster Assignment} step from Algorithm \ref{alg: cluster} using the centroids $C_1,\ldots,C_k$ without updating the clusters any further. This yields clusters $\mcal{Y}_1,\ldots,\mcal{Y}_k$ such that $\mcal{Y}_i$ and $\mcal{X}_i$ share the same centroid for each $i$. Now fix $\lambda\in[0,1]$ and compute

\begin{equation*}
\mcal{Z}_i=g\big(\lambda h(\pc_i)+(1-\lambda) h(\mcal{Y}_i)\big)
\end{equation*}
for each $i=1,\ldots,k$ and set $\mcal{Z}=\cup_{i=1}^k \mcal{Z}_i$.
Then $\mcal{Z}$ is a new point cloud which mixes the geometry of the point clouds $\pc$ and $\mcal{Y}$. The larger $\lambda$ the more it will maintain properties of $\pc$, the smaller $\lambda$ the more $\mcal{Z}$ will tend towards $\mcal{Y}$.

\item \label{pipe: styletransfer}\textit{Transfering a style onto a point cloud}. Let $\pc$ denote a point cloud and $\mcal{S}$ a design point cloud, see Subsection \ref{subsec: styletransfer}. Again by normalizing we assume that both lie within the cube $[0,1]^3$. As discussed in Subsection \ref{subsec: styletransfer} we compute $\mcal{S}_{\pc}$ and apply \ref{pipe: blending} where $\mcal{Y}=\mcal{S}_{\pc}$.
\end{enumerate}

\begin{center}
\begin{figure}[htbp]
\includegraphics[width=0.9\textwidth]{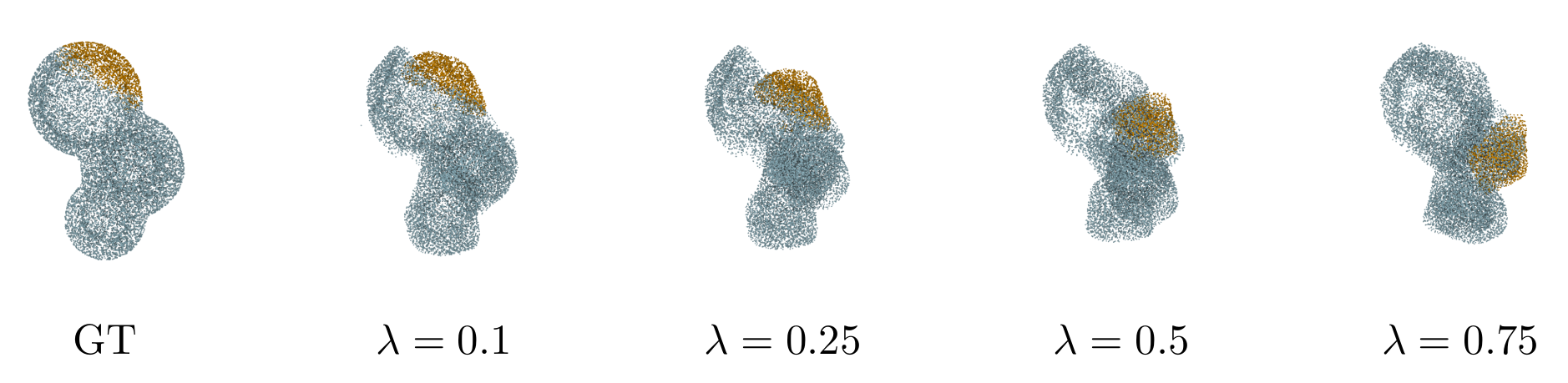}
\caption{Blending process between a point cloud and itself, where the clusters are not accordingly matched, c.f.\ Remark \ref{rem::badclusters}.}
\label{fig::badclusters}
\end{figure}
\end{center}

\begin{remark}\label{rem::badclusters}

Note that by using the same clusters for $\mcal{X}$ and $\mcal{Y}$ we avoid possible translations that might occur during the blending process between the clusters.
Indeed, in Figure \ref{fig::badclusters} we see a blending between a point cloud and itself, i.e.\ $\mcal{X}=\mcal{Y}$, but instead of using the reassembling method discussed in \ref{pipe: blending}, $\mcal{X}$ and $\mcal{Y}$ are clustered independently resulting into two different sets of centroids.
We then naively match the clusters by fixing a cluster of $\mcal{X}$ and choose a cluster of $\mcal{Y}$ where the centroids have minimal distance and repeat this procedure for every cluster.
But since each cluster can only be used once, there might emerge matchings, which are not suitable. This is also seen in Figure \ref{fig::badclusters}. On the left we see the ground truth (GT) and one cluster in orange at the top of the point cloud. Due to bad matching, we see in Figure \ref{fig::badclusters} that this orange cluster is translated to a cluster near the bottom of the point cloud, which essentially ruins the obtained point clouds $\mathcal{Z}_i$ for $1<i<k$, using the notation in \ref{pipe: blending}. A similar behaviour can also be observed in Figure \ref{fig::clustercomparison} on a larger scale.
\end{remark}

%%%%%%%%%%%%%%%%%%%%%%%%%%%%%%
% Bad clusters

\section{Dataset}
\label{sec: dataset}
\begin{figure}[htbp]
\centering
\includegraphics[width=0.9\textwidth]{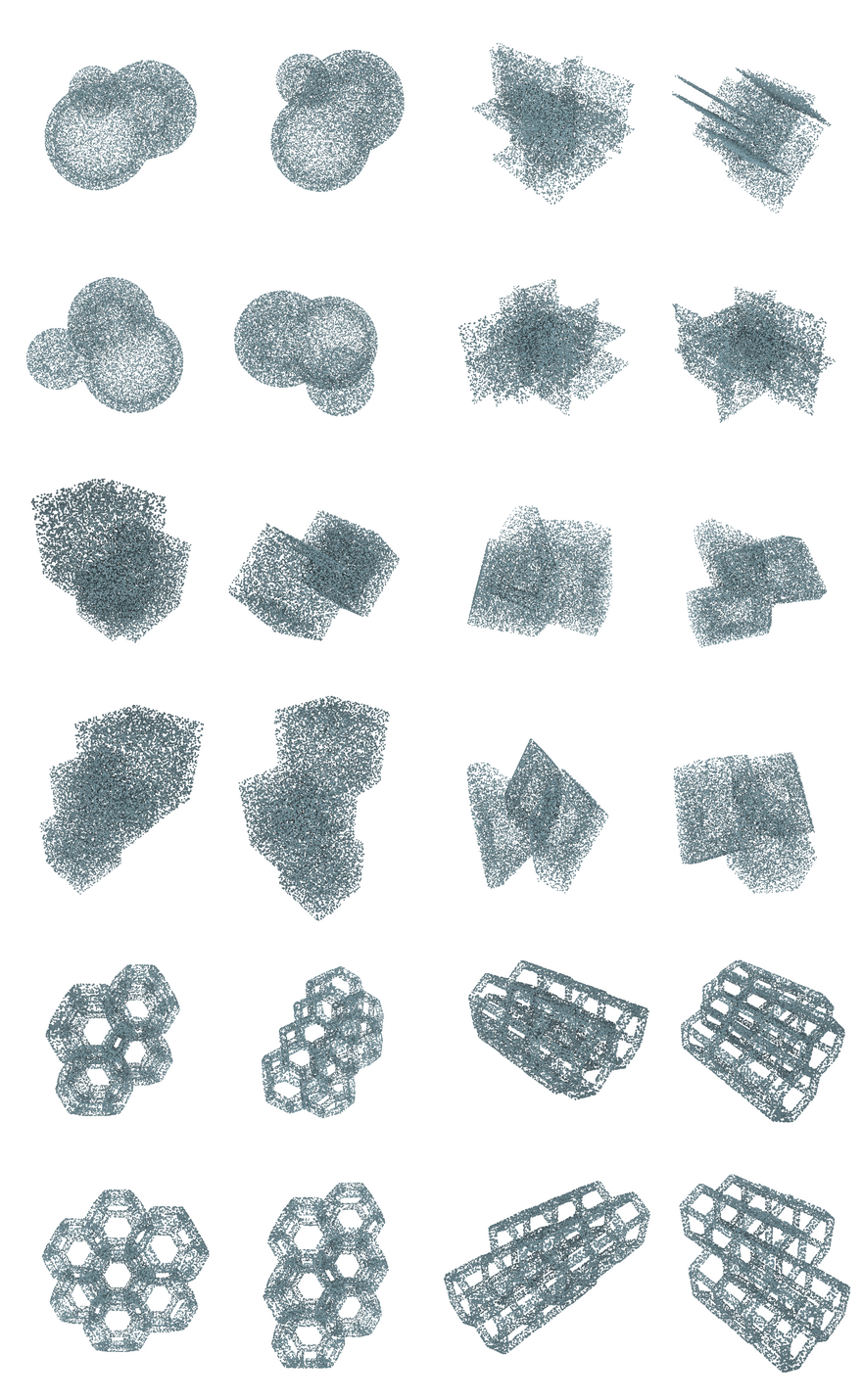}
\caption{Six point clouds from our training dataset. Each point cloud is shown from four different angles.} \label{fig: trainingdata}
\end{figure}

\begin{figure}[htbp]
\centering
\includegraphics[width=0.5\textwidth]{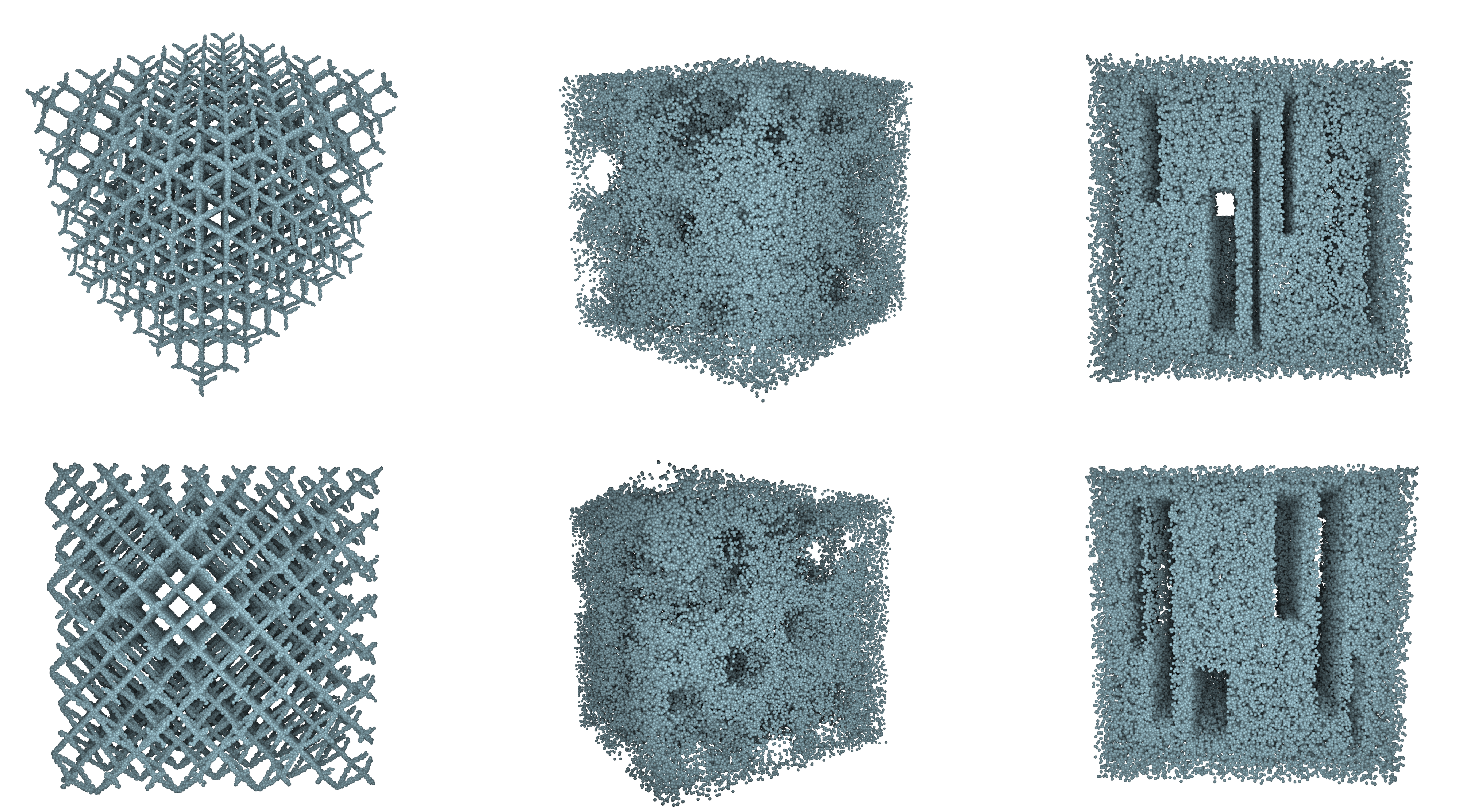}
\caption{Our three design point clouds. Left: Stripes; Center: Porous; Right: Cuts.}\label{fig: designcubes}
\end{figure}

While the original autoencoder in \cite{foldingnet} was trained on the Shapenet dataset \cite{shapenet}, we trained our network on a new training set $\mcal{D}$, which allows the autoencoder to reconstruct volumetric point clouds.
Our dataset contains 2107 point clouds, each consisting of 16384 points. Essentially the training dataset can be categorized into four classes: encapsulated spheres, encapsulated cuboids, orthogonally intersecting planes and lattices with various geometric shapes, see Figure \ref{fig: trainingdata}.
Furthermore we provided three design point clouds $\mcal{S}_1,\mcal{S}_2,\mcal{S}_3$ representing a \textit{stripes}, \textit{porous} and \textit{cut} design, respectively, see Figure \ref{fig: designcubes}.
For each $\pc\in\mcal{D}$ we additionally computed the corresponding $\pc_{\mcal{S}_1},\pc_{\mcal{S}_2},\pc_{\mcal{S}_3}$, see Subsection \ref{subsec: styletransfer}. Since we want to apply the autoencoder to high resolution point clouds, we then split each point cloud $\pc$ in our training set $\mcal{D}$ as well as the resulting designs $\pc_{\mcal{S}_1},\pc_{\mcal{S}_2},\pc_{\mcal{S}_3}$ into clusters of equal size, as discussed in Subsection \ref{subsec: clustering}, and used the resulting clusters for training.
We will denote this set of clusters also by $\mcal{D}$. As we will discuss in Section \ref{sec: exp setup and results}, this dataset is rich enough in geometric properties in order for the autoencoder to learn how to reconstruct and blend between various geometric structures as well as transfer one of the design point clouds $\mcal{S}_1,\mcal{S}_2$ and $\mcal{S}_3$ onto new point clouds.
%%%%%%%%%%%%%%%%%%%%%%%%%%%%%%%%%%%%%%%%%
% Training data

%%%%%%%%%%%%%%%%%%%%%%%%%%%%%%%%%%%%%
% Design pcs

\section{Experimental Setup and Results}
\label{sec: exp setup and results}
All our experiments have been implemented using the PyTorch library \cite{pytorch}. For our experiments we trained the network on clusters of size 2048, i.e.\ $m=n=2048$ and used a feature space dimension of $d=512$. The hyperparameter for the nearest neighbour approach in the graph layers was set to $k=16$, while the weights for the loss were set to 
\[
\alpha_{\text{EMD}}=1/\max_{\pc,\mcal{Y}\in\mcal{D}}\mathrm{EMD}(\pc,\mcal{Y})
\]
and similarly
\[
\alpha_{\text{Chamfer}}=1/\max_{\pc,\mcal{Y}\in\mcal{D}}\mathrm{Chamfer}(\pc,\mcal{Y})\,,
\]
where in both cases the maximum was estimated by using a Monte-Carlo method. We calculate the Sinkhorn divergence in order to approximate the EMD, using the GeomLoss library \cite{geomloss}. Its blur is set to $10^{-3}$ and the scaling is 0.9.  The learning rate was $10^{-3}$ and the network was trained for 300 epochs. For the style transfer method described in \ref{pipe: styletransfer}, the density is estimated using a Gaussian kernel density estimater with bandwidth 0.01. When sampling the point cloud $\pc_{\mcal{S}}$ note that depending on $\mcal{X}$ and $\mcal{S}$ there might be areas where the density is 0, thus possibly resulting in too few points to sample from the design $\mcal{S}$. We solve this by sampling points with replacement and perturb them by adding $0.1\%$ Gaussian white noise to each point. 
The code for this can be found at \url{https://github.com/antholzer/pointcloudmatcher}.

\begin{figure}[htbp]
\centering
\includegraphics[width=0.5\textwidth]{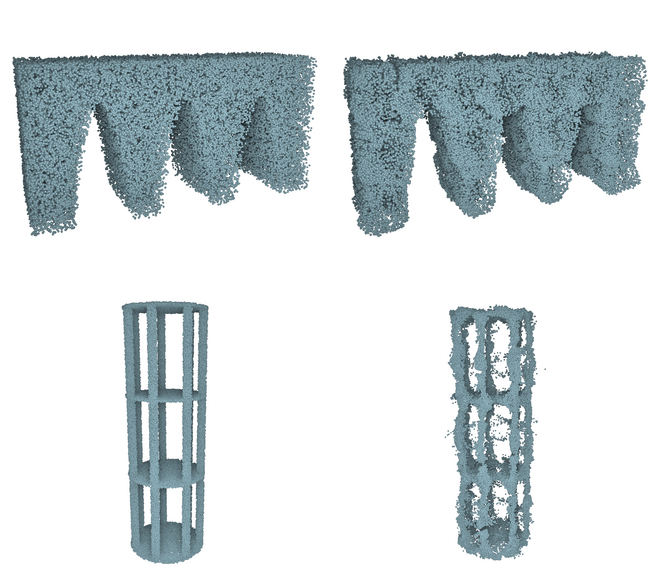}
\caption{Reconstructions of two point clouds representing a bridge and pillar, respectively. Left shows the ground truth, right the reconstruction using our autoencoder.}
\label{fig::reconstruction}
\end{figure}

In Figure \ref{fig::reconstruction} we see on the left-hand side two point clouds representing a bridge and pillar respectively as ground truths. Both point clouds consist of 81920 points and were not part of the training data. On the right-hand side we see the reconstruction of the autoencoder. In both cases the reconstructions maintain the shape as well as interior structure. Even the fragile structure of the pillar point cloud is reproduced.

%%%%%%%%%%%%%%%%%%%%%%%%%%%%%%%%%%%
% Reconstruction results

\begin{figure}[htbp]
\begin{center}
\includegraphics[width=0.9\textwidth]{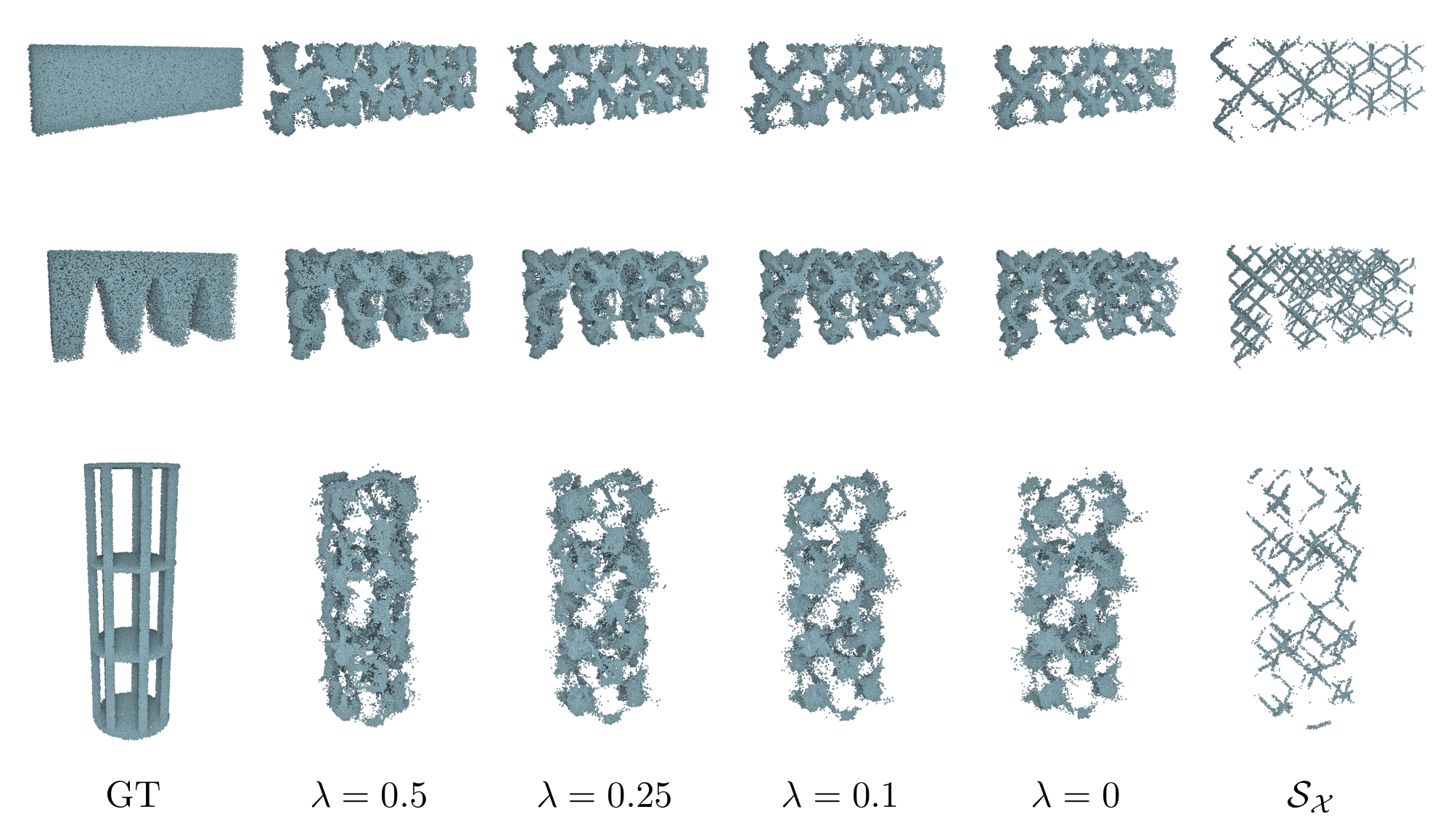}
\caption{The stripes design applied to a wall (top), bridge (center) and pillar (bottom).}\label{fig::transfer_stripes}
\end{center}
\end{figure}

\begin{figure}[htbp]
\begin{center}
\includegraphics[width=0.9\textwidth]{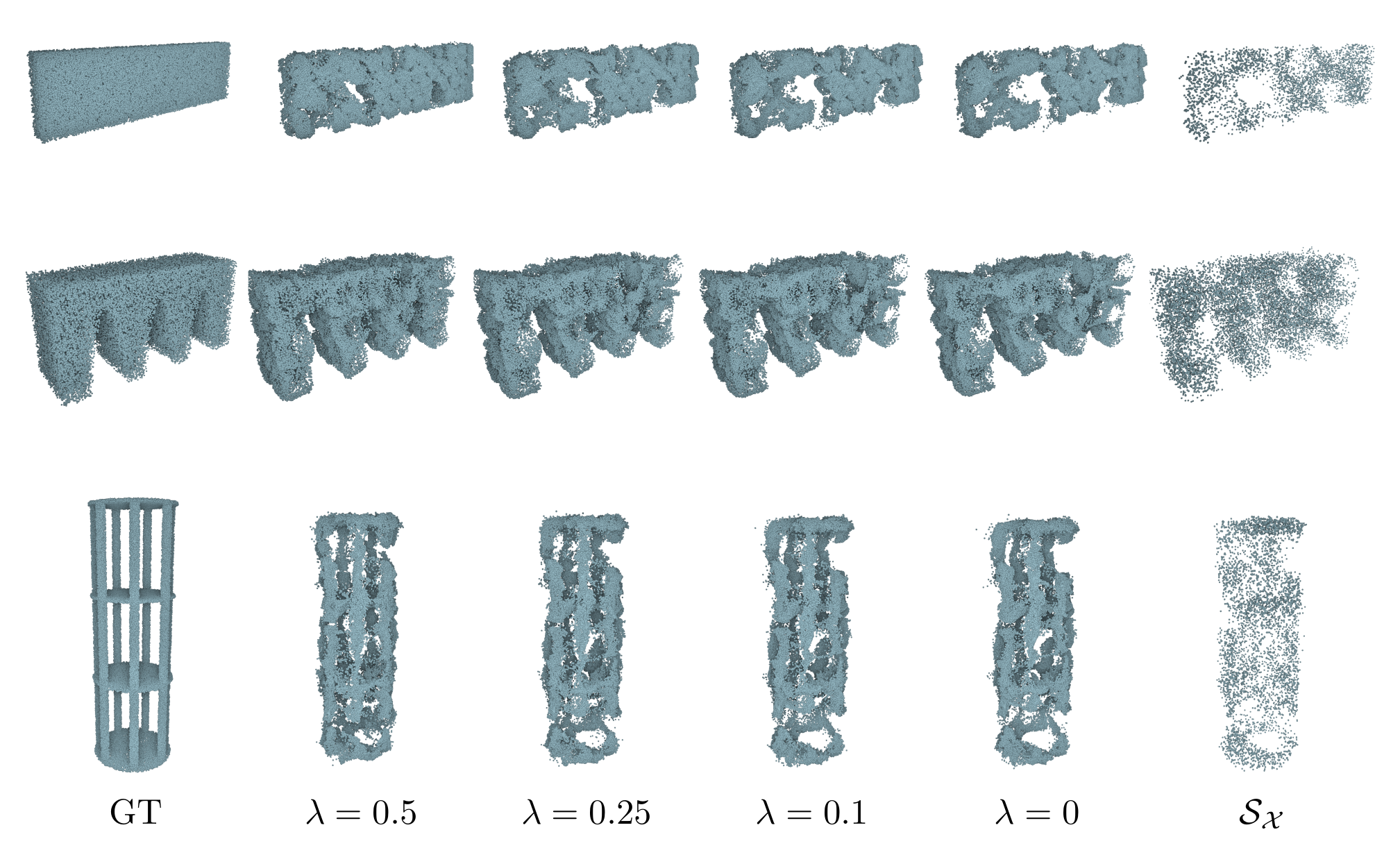}
\caption{The porous design applied to a wall (top), bridge (center) and pillar (bottom).}\label{fig::transfer_porous}
\end{center}
\end{figure}

\begin{figure}[htbp]
\begin{center}
\includegraphics[width=0.9\textwidth]{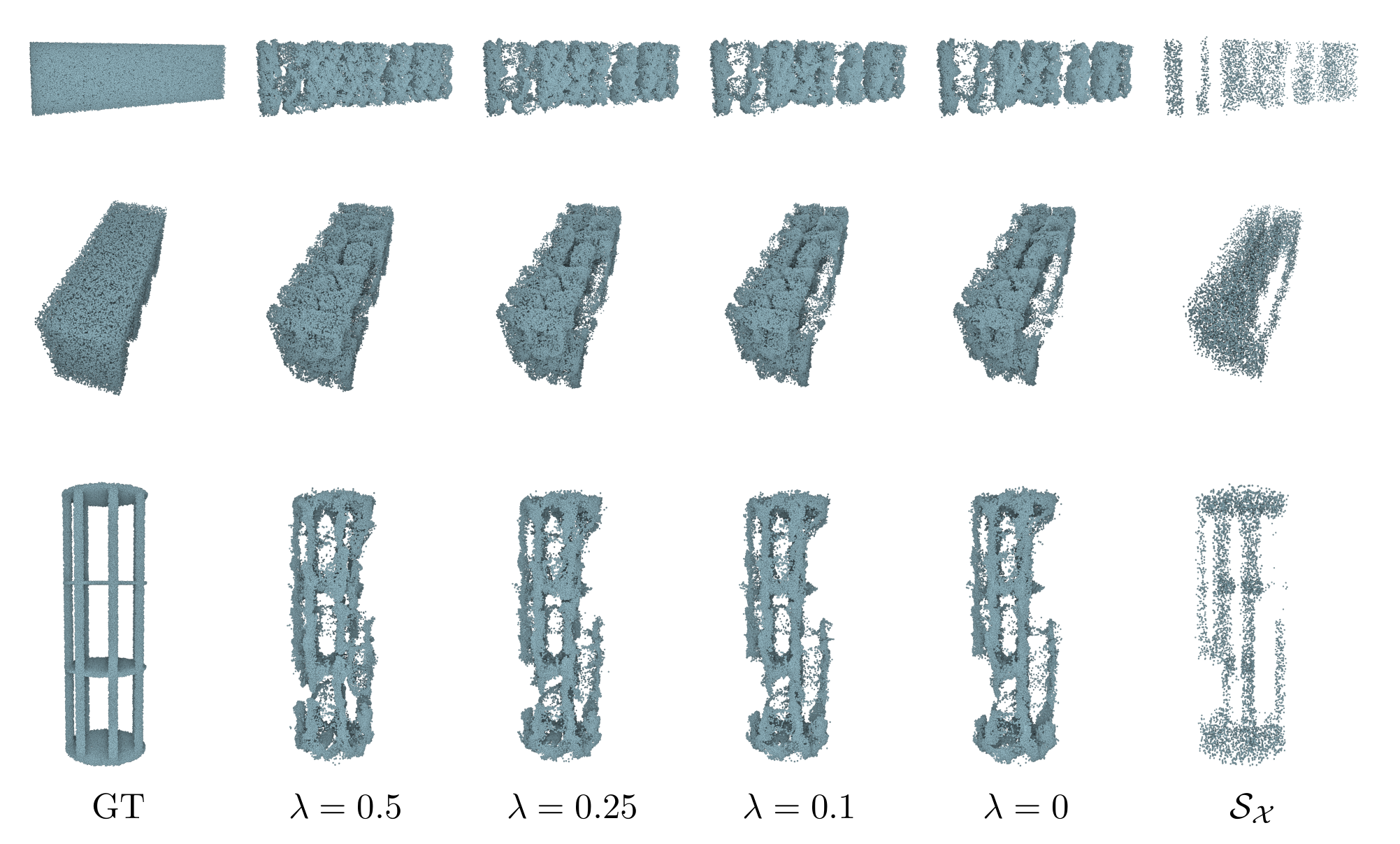}
\end{center}
\caption{The cut design applied to a wall (top), bridge (center) and pillar (bottom).}\label{fig::transfer_cut}
\end{figure}

Let us discuss our results concerning the style transfer from \ref{pipe: styletransfer}. In the Figures \ref{fig::transfer_stripes}, \ref{fig::transfer_porous} and \ref{fig::transfer_cut} we see the results when applying the stripes, porous and rectangular design, respectively, to a point cloud representing a wall (top), also consisting of 81920 points, and the already mentioned bridge (middle) and pillar (bottom). First note that in each case the sampled point cloud $\mcal{S}_\mcal{X}$ consists of very thin lines/structures. This is due to the sampling with replacement method and the little noise we add. The higher the noise to signal ratio, the thicker but also frayed $\mcal{S}_\mcal{X}$ would become. This could result in $\mcal{S}_{\pc}$ loosing essential design properties of $\mcal{S}$. However, as shown in our results, the constructed $\mcal{S}_{\pc}$ are sufficient to generate point clouds which maintain the original geometric shape of $\mcal{X}$, while imposing the design structure of $\mcal{S}$. Indeed in Figure \ref{fig::transfer_stripes} the ``zick zack'' design is inhereted from the point clouds and reduces depending on the factor $\lambda$. Simultaneously each point cloud maintains the shape of the ground truth. While in the first line of Figure \ref{fig::transfer_stripes} we see the style transfer process between a simple point cloud representing a wall, we observe in the second and third line, that the shape is even maintained for more complex structures like a bridge and a mutilayered pillar. When inspecting the style transfers regarding the porous and cut design in Figure \ref{fig::transfer_porous} and \ref{fig::transfer_cut} respectively, we essentially observe the same outcome. The volumetric design is applied on the ground truths while maintaining their geometric properties depending on the parameter $\lambda$. Note though that one downside we observe is that the point clouds tend to frazzle, which makes especially the edges of the maintained point clouds not as precise as desired. However, this could be solved by denoising the point clouds in a post processing step. On a final note, observe especially in each example the case $\lambda=0$ where we see the reconstruction of the corresponding $\mcal{S}_{\pc}$. In every case the reconstruction captures voids as well as the shape while thickening the structures, which already resembles the ground truth.
%%%%%%%%%%%%%%%%%%%%%%%%%%%%%%%%%%%%
% Stripes Style Transfer

%%%%%%%%%%%%%%%%%%%%%%%%%%%%%%%%%%%%
% Porous Style Transfer
%
% %%%%%%%%%%%%%%%%%%%%%%%%%%%%%%%%%%%%
% % Cut Style Transfer

\section{Conclusion}
\label{sec: Conclusion}
Let us summarize our results. In order to obtain real-world applicable autoencoders for point clouds, we discussed a cluster and reassembling method which allows to subdivide a high resolution point cloud into smaller point clouds and apply the autoencoder cluster wise. As an example of our method we additionally proposed an adaption of the autoencoder in \cite{foldingnet} to volumetric point clouds.  We then described how these two procedures can be combined to obtain a density-based style transfer method for volumetric point clouds. Our results clearly show the effectiveness of our cluster and reassembling methods as well as the style transfer strategy.

\section{Acknowledgements}
This project is funded by the Austrian Science Fund (FWF): P8470-029-011. We thank Mathias Bank, Tilman Fabini and Viki Sandor for generating the training dataset. 
\bibliography{ref.bib}
\end{document}